\documentclass[10pt,twocolumn,letterpaper]{article}

\usepackage[pagenumbers]{cvpr} 

\usepackage{graphicx}       
\usepackage{xspace}         
\usepackage{color}
\usepackage{xcolor}
\usepackage{colortbl}
\usepackage{amsfonts,bm}
\usepackage{multirow}
\usepackage{tabularx}
\usepackage{threeparttable}
\usepackage{upgreek}
\usepackage{pifont}
\usepackage{bbding}
\usepackage{makecell}
\usepackage{wrapfig}
\usepackage{soul}
\usepackage{wasysym}
\usepackage{subcaption}  
\usepackage[first=0,last=9]{lcg}
\usepackage{slashbox}
\usepackage[ruled,vlined]{algorithm2e}  

\definecolor{Gray}{gray}{0.85}

\newcolumntype{x}[1]{>{\centering\arraybackslash}p{#1pt}}
\newcolumntype{y}[1]{>{\raggedright\arraybackslash}p{#1pt}}
\newcolumntype{z}[1]{>{\raggedleft\arraybackslash}p{#1pt}}

\newcommand{\fullname}{Adaptive Keyframe Sampling\xspace}
\newcommand{\name}{AKS\xspace}

\definecolor{cvprblue}{rgb}{0.21,0.49,0.74}
\usepackage[pagebackref,breaklinks,colorlinks,citecolor=cvprblue]{hyperref}

\title{\fullname for Long Video Understanding}

\author{Xi Tang\textsuperscript{1}\thanks{Equal contribution.}, ~Jihao Qiu\textsuperscript{1}\footnotemark[1] ,  ~Lingxi Xie, ~ Yunjie Tian\textsuperscript{2}, ~Jianbin Jiao\textsuperscript{1}, ~Qixiang Ye\textsuperscript{1}\\
\textsuperscript{1}University of Chinese Academy of Sciences\\ \quad 
\textsuperscript{2}University at Buffalo, SUNY\\
{\tt\small \{tangxi19,qiujihao19\}@mails.ucas.ac.cn}\quad{\tt\small \{jiaojb,qxye\}@ucas.ac.cn}
}

\begin{document}

\maketitle

\begin{abstract}
Multimodal large language models (MLLMs) have enabled open-world visual understanding by injecting visual input as extra tokens into large language models (LLMs) as contexts. However, when the visual input changes from a single image to a long video, the above paradigm encounters difficulty because the vast amount of video tokens has significantly exceeded the maximal capacity of MLLMs. Therefore, existing video-based MLLMs are mostly established upon sampling a small portion of tokens from input data, which can cause key information to be lost and thus produce incorrect answers.
This paper presents a simple yet effective algorithm named \fullname (\textbf{\name}). It inserts a plug-and-play module known as keyframe selection, which aims to maximize the useful information with a fixed number of video tokens. We formulate keyframe selection as an optimization involving (1) the \textbf{relevance} between the keyframes and the prompt, and (2) the \textbf{coverage} of the keyframes over the video, and present an adaptive algorithm to approximate the best solution. Experiments on two long video understanding benchmarks validate that \name improves video QA accuracy (beyond strong baselines) upon selecting informative keyframes. Our study reveals the importance of information pre-filtering in video-based MLLMs.Our codes are available at \href{https://github.com/ncTimTang/AKS}{https://github.com/ncTimTang/AKS}.

\end{abstract}

\section{Introduction}
\label{sec:introduction}

\begin{figure}[t]
\centering
\includegraphics[width=0.99\linewidth]{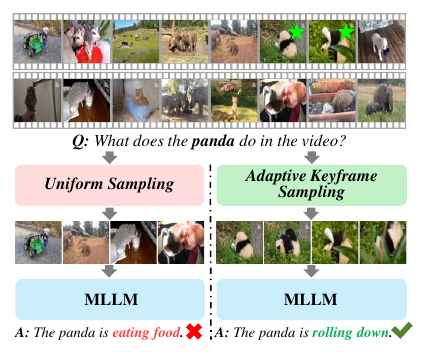}
\caption{The accuracy of video-based MLLMs heavily relies on the quality of keyframes. The above example shows a long video from VideoMME~\cite{fu2024video} where keyframes are marked with green stars. The same MLLM (\textit{i.e.}, LLaVA-Video~\cite{zhang2024video}) is used for answering the question. Uniform sampling (the default setting in~\cite{zhang2024video}) finds irrelevant frames (the MLLM mostly performs a random guess), while our algorithm (\name) finds keyframes and produces the correct answer.}
\label{fig:intro}
\end{figure}

\qquad\quad``\textit{You can't manage what you can't measure.}''

\hfill\textit{— Peter Drucker}

\noindent
Recent years have witnessed a rapid development of multimodal large language models (MLLMs)~\cite{liuVisualInstructionTuning2023,tian2024chatterbox,zhangGPT4RoIInstructionTuning2023,laiLISAReasoningSegmentation2023} for open-world visual understanding. Among the large corpus of research in MLLMs, a straightforward and important direction is to generalize them to video data. Compared to still images, videos contain richer and more complex visual content, thus raising serious challenges to MLLMs including key information retrieval, summarization, logical inference, \textit{etc}. Many benchmarks~\cite{wu2024longvideobench,fu2024video} have been established to evaluate MLLMs for video understanding.

A typical framework of image-based MLLMs involves encoding the input image into a set of visual tokens and feeding them as the context of LLMs. When this framework was transplanted to videos, especially long videos, a difficulty arose from the limited capacity of MLLMs, \textit{i.e.}, the maximal number of visual tokens that MLLMs can process is much fewer than that of an entire video; in other words, not all video tokens can be perceived by MLLMs. To bridge the gap, recent approaches~\cite{lin2023videollava,xu2024pllava} often sampled a small portion of frames from the input video; consequently, the performance of these MLLMs heavily relies on the quality of selected frames (\textit{i.e.}, keyframes). Despite its importance, the keyframe selection algorithm has not been carefully designed, \textit{e.g.}, LLaVA-Video~\cite{zhang2024video} simply applied a uniform sampling strategy which, as shown in Figure~\ref{fig:intro}, is prone to losing important information and thus leads to incorrect outputs of video understanding.

This paper presents a systematic study on keyframe selection and reveals its importance to video understanding and beyond. We formulate keyframe selection as a plug-and-play module before the MLLM's visual encoder; its goal is to maximize the usefulness of the keyframes in video understanding. Intuitively, we propose two key aspects to be considered, namely, (1) \textbf{relevance} (\textit{i.e.}, how the keyframes are related to the question) and (2) \textbf{coverage} (\textit{i.e.}, how the keyframe set covers the useful information in the entire video). Specifically, we quantify the target by (1) computing relevance between each candidate frame and the prompt using a vision-language (VL) model, and (2) estimating coverage by recursively partitioning the video into bins and counting the number of keyframes within each bin. We show that maximizing relevance and coverage alone produces simple baselines for keyframe selection, while a proper tradeoff between them, obtained by the proposed \fullname (\textbf{\name}) algorithm, leads to the best practice of video understanding.

We evaluate our approach on LongVideoBench~\cite{wu2024longvideobench} and VideoMME~\cite{fu2024video}, two benchmarks for long video understanding. We investigate three frame-based MLLMs (Qwen2VL~\cite{wang2024qwen2}, LLaVA-OV~\cite{li2024llava}, and LLaVA-Video~\cite{zhang2024video}) as the baseline and insert \name as an off-the-shelf module to improve the quality of keyframes. Our approach achieves consistent accuracy gain throughout all tests. Specifically, when \name is integrated with LLaVA-Video-7B, we set new records on these two benchmarks with 7B models. We further validate that the improvement owes to higher-quality keyframes found by \name, demonstrating that MLLMs become stronger with more informative visual contexts. Our study reveals that pre-filtering visual data is crucial and will be a long-lasting research topic for MLLMs in perceiving high-dimensional data, \textit{e.g.}, long videos, and even 4D data.

\section{Related Work}
\label{sec:related_work}

\noindent\textbf{Large language models (LLMs) and multimodal LLMs (MLLMs).}
LLMs~\cite{devlin2018bert,brown2020language,chung2022scaling,thoppilan2022lamda,chowdhery2022palm,zhang2022opt,touvron2023llama,zeng2022glm,chiang2023vicuna} have marked a new era in AI, showcasing significant potential in unifying various tasks covering language understanding and generation.
To extend LLMs for visual understanding, the community has focused on aligning visual and language data within a unified feature space~\cite{radford2021learning}. There are generally two types of approaches, (1) internal adaptation, such as~\cite{alayrac2022flamingo}, that integrates cross-attention mechanisms within LLMs to achieve vision-language alignment, and (2) external adaptation, such as~\cite{liBLIP2BootstrappingLanguageImage2023,instructblip,liuVisualInstructionTuning2023}, that trains additional modules for the same purpose. 
Consequently, vision foundation models~\cite{dosovitskiy2020image,liu2021swin,radford2021learning,tianIntegrallyPreTrainedTransformer2023,tian2021semantic,zhang2022hivit,kirillovSegmentAnything2023} have evolved into multimodal LLMs (MLLMs)~\cite{liuVisualInstructionTuning2023,tian2024chatterbox,zhangGPT4RoIInstructionTuning2023,laiLISAReasoningSegmentation2023}, enabling them to perform language-guided visual understanding tasks.

\noindent\textbf{Video-based MLLMs.}
Researchers have extended MLLMs to video understanding. Early efforts in this area include VideoChat~\cite{li2024videochat}, Video-ChatGPT~\cite{maaz2023videochatgpt}, Video-LLaMA~\cite{zhang2023videollama}, Video-LLaVA~\cite{lin2023videollava}, LanguageBind~\cite{zhu2024languagebind}, and Valley~\cite{luo2023valley}, \textit{etc}. Different from still images, videos contain rich content that, when encoded as visual tokens, exceed the maximal context capacity of MLLMs. Most of the above methods have sampled video frames to fit MLLMs; some of them, such as Video-ChatGPT~\cite{maaz2023videochatgpt}, introduced more efficient video features. There are also studies on instance-level video understanding have been proposed, such as LEGO~\cite{li2024groundinggptlanguage} for moment retrieval, PG-Video-LLaVA~\cite{munasinghe2023pgvideollava} for video grounding, and Artemis~\cite{qiu2024artemis} for video referring, enriching the corpus of video understanding.

\noindent\textbf{MLLMs for Long Video Understanding.}
Going one step further, long video understanding faces greater challenges due to the increased difficulty of keyframe selection, leading to significant loss of critical information. While some MLLMs (\textit{e.g.}, Kangaroo~\cite{liu2024kangaroo} and LLaVA-Video~\cite{zhang2024video}) utilize language models with larger context capacities to allow more frames to be encoded and processed, many others have designed specific strategies to mitigate this issue.
For example, MovieChat~\cite{song2024moviechat} employed both short-term and long-term memory banks to compress and preserve video content. Similarly, MA-LMM~\cite{he2024ma} VideoStreaming~\cite{qian2024streaming} used a Q-former and a small language model (phi-2~\cite{javaheripi2023phi}), to condense video data, while LongVLM~\cite{weng2024longvlm} adopted token merging to decrease the number of video tokens. Goldfish~\cite{ataallah2024goldfish} integrated short video understanding with information retrieval to answer complex queries. In summary, these approaches aim to reduce the number of video tokens, but there is often no guarantee that key information in the video can be preserved. \textit{This work presents a simple yet effective algorithm that maximally preserves important information for long video understanding.}

\begin{figure*}[t]
\vspace{-0.7cm}
\centering
\includegraphics[width=0.99\linewidth]{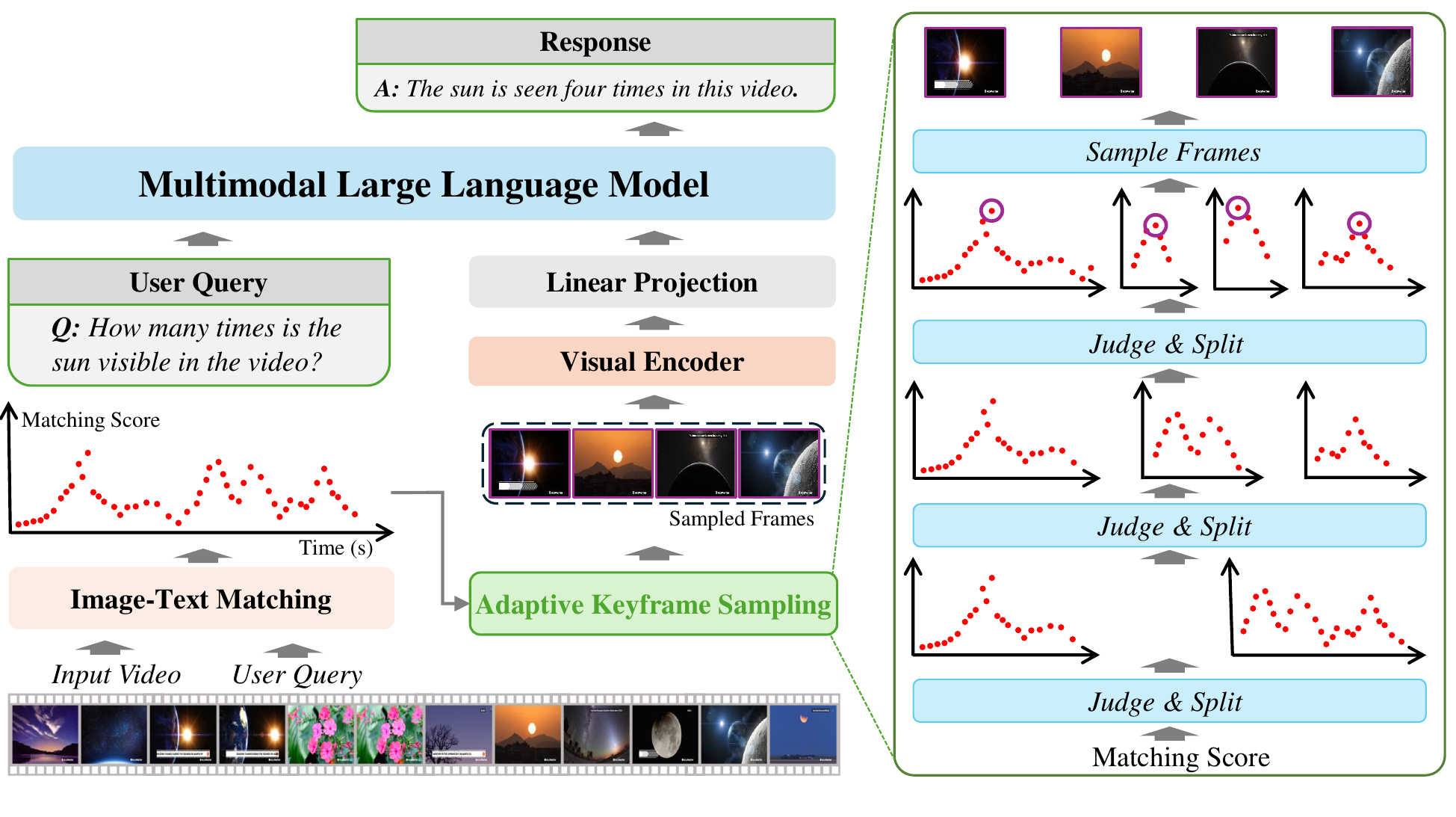}
\caption{The overall framework of our approach. We insert a plug-and-play module, \fullname (\name, marked in green frames) into the MLLM to improve the quality of sampled keyframes. Each red dot indicates a prompt-frame matching score (\textit{i.e.}, $s(\mathbf{Q},\mathbf{F}_t)$, see Section~\ref{method:principles}). \name follows a recursive, judge-and-split optimization for keyframe selection (see Section~\ref{method:optimization}).}
\label{fig:method}
\end{figure*}

\section{Method}
\label{sec:method}

\subsection{Preliminaries}
\label{method:preliminaries}

Within a broad range of video understanding tasks, the model receives a video clip and a text instruction as input and is required to output a text answer. Without loss of generality, we denote the video as $\mathbf{V}\in\mathbb{R}^{T\times W\times H\times C}$, where $T$ is the number of frames and $W$, $H$, and $C$ denote the width, height, and number of channels, respectively. We consider each frame an image $\mathbf{V}_t$ ($t\in\{1,2,\ldots,T\}$) and apply a pre-trained encoder (\textit{e.g.}, the CLIP ViT-L model~\cite{radford2021learning}) to extract a set of visual tokens $\mathbf{F}_t$ from it. The text instruction (\textit{a.k.a.}, prompt) is denoted as $\mathbf{Q}$.

The overall pipeline of our algorithm is illustrated in Figure~\ref{fig:method}. We use a regular MLLM that addresses video understanding with the template \textsf{[ \textcolor{blue}{\textbf{User:}} $\langle$video-tokens$\rangle$ $\langle$text-instruction$\rangle$ \textcolor{blue}{\textbf{Assistant:}} ]}, where the video tokens and text instruction are projected into the same feature space using an MLP. For simplicity, we denote the MLLM as a function of $G(\{\mathbf{F}_t\})$ where we omit the LLM part and only focus on the visual tokens as contexts. With a limited capacity of visual contexts (\textit{i.e.}, the number of video tokens cannot exceed a specific value), the above pipeline encounters difficulty in dealing with long videos where not all video content can be perceived by the MLLM.

A straightforward solution is to select \textbf{keyframes} from the input video for token extraction. In other words, the goal is to design a selection function $\mathrm{KS}_M(\mathbf{Q},\mathbf{F})$ that outputs an index set, $\mathcal{I}\subseteq\{1,2,\ldots,T\}$ and $|\mathcal{I}|=M$, indicating the $M$ best keyframes ($M$ is pre-defined according to the context capacity of the MLLM). Video tokens extracted from the keyframes (\textit{i.e.}, $\{\mathbf{F}_t\mid t\in\mathrm{KS}_M(\mathbf{Q},\mathbf{F})\}$ compose of the context of the MLLM. As shown in Figure~\ref{fig:intro}, the quality of keyframe selection is crucial for video understanding, but the function $\mathrm{KS}_M(\mathbf{Q},\mathbf{F})$ has not been well studied in the community. As an example, a recent MLLM for video understanding~\cite{zhang2024video} simply performed uniform sampling for keyframe selection; with the function $\mathrm{KS}_M(\cdot)$ not using $\mathbf{Q}$ and $\mathbf{F}$ at all, it cannot guarantee to find useful information for question answering.

In what follows, we establish two principles of keyframe selection (Section~\ref{method:principles}), after which we will present \name, our optimization algorithm (Section~\ref{method:optimization}).

\begin{figure*}[!t]
\vspace{-0.7cm}
\centering
\includegraphics[width=0.95\linewidth]{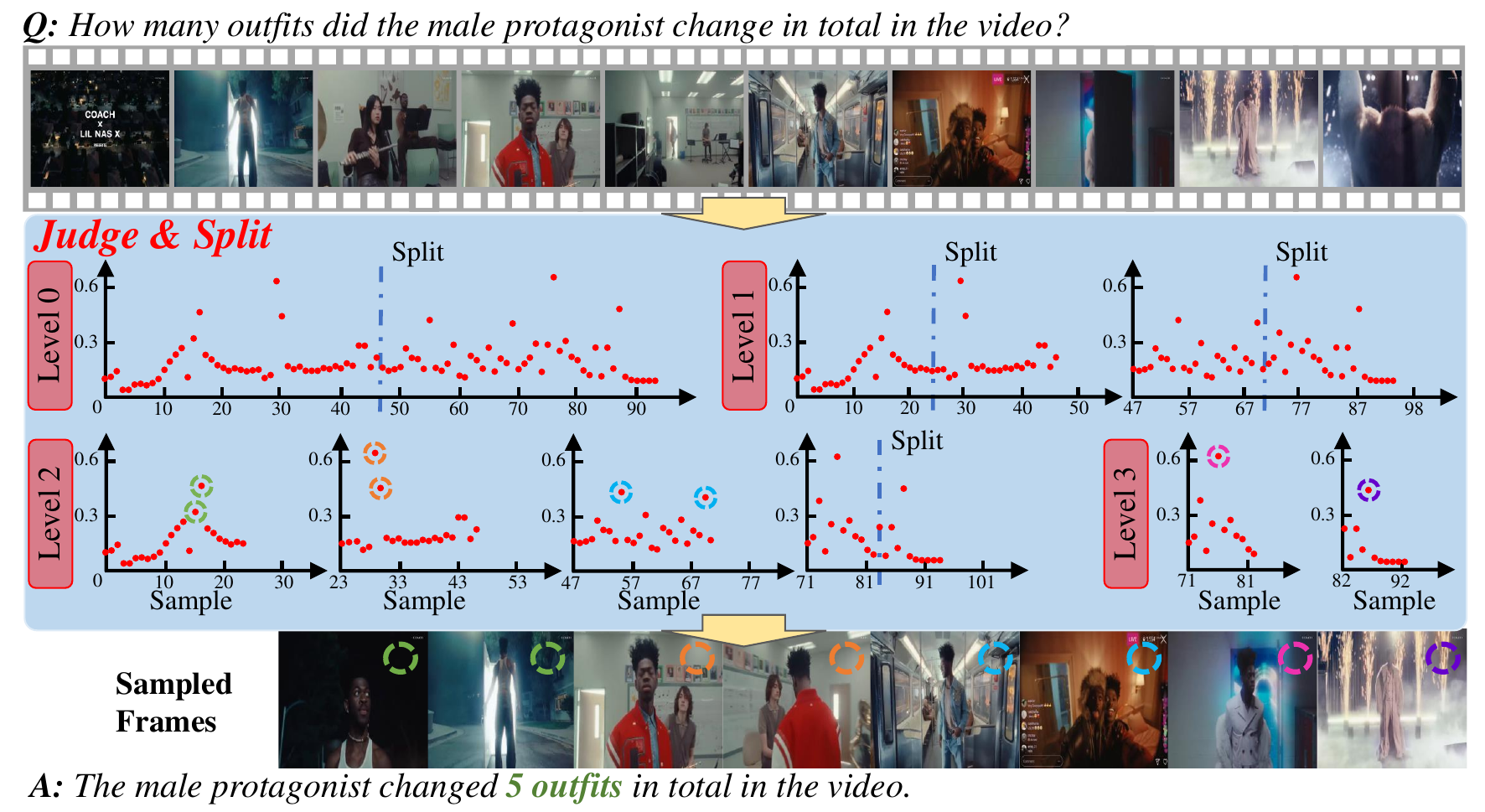}
\caption{An example of adaptive sampling (\textbf{ADA}). $8$ keyframes are to be selected from the input video. Each red dot indicates a prompt-frame matching score, $s(\mathbf{Q},\mathbf{F}_t)$. At Level-$0$ and  Level-$1$, all bins are \textbf{split} into two sub-bins; at Level-$2$, only the rightmost bin is further partitioned while the top-$2$ scores are \textbf{sampled} from the other three bins. Level-$3$ has reached the maximal depth.}
\label{fig:split}
\end{figure*}

\subsection{Principles of Keyframe Selection}
\label{method:principles}

The keyframe selection function $\mathrm{KS}_M(\mathbf{Q},\mathbf{F})$ is to maximize the amount of useful information, \textit{i.e.},
\begin{equation}
\label{eqn:objective1}
\mathrm{KS}_M(\mathbf{Q},\mathbf{F})=\arg\max_{|\mathcal{I}|=M}G'(\{\mathbf{F}_t\mid t\in\mathcal{I}\}).
\end{equation}
Here, we assume $G'(\cdot)$ to be a complementary function of $G(\cdot)$, indicating the MLLM's confidence about its output. Eqn~\eqref{eqn:objective1} is mathematically intractable due to two reasons. First, the optimization involves exponentially many candidates of $\mathcal{I}$. Second and more importantly, the function $G'(\cdot)$ is difficult to estimate because there is no supervision for keyframe selection -- even when a training set is available and one can compare the output of $G(\cdot)$ with the ground-truth answer, it is not guaranteed that a correct answer corresponds to a perfect set of keyframes, and vice versa.

We propose a heuristic method to approximate Eqn~\eqref{eqn:objective1}. Intuitively, a set of keyframes is informative when the following conditions are satisfied. (1) The \textbf{relevance} between each frame and the prompt is high, \textit{i.e.}, the visual data is useful for question answering. (2) The \textbf{coverage} of the selected frames is sufficient to comprehensively answer the question. Note that the coverage is difficult to quantify, and the second principle is closely related to preventing redundant frames (\textit{e.g.}, neighboring frames with almost the same visual content) from being selected, because (when the size of $|\mathcal{I}|$ is fixed) they can potentially reduce the amount of other useful information and thus harm the coverage of the entire keyframe set.

Following the analysis above, we reformulate the right-hand side of Eqn~\eqref{eqn:objective1}, yielding:
\begin{equation}
\label{eqn:objective2}
\mathrm{KS}_M(\mathbf{Q},\mathbf{F})=\arg\max_{|\mathcal{I}|=M}\sum_{t\in\mathcal{I}}s(\mathbf{Q},\mathbf{F}_t)+\lambda\cdot c(\mathcal{I}).
\end{equation}
Here we introduce two quantities, $r(\mathbf{Q},\mathbf{F}_t)$ as the relationship between the prompt $\mathbf{Q}$ and the $t$-th frame $\mathbf{F}_t$, and $c(\mathcal{I})$ as the coverage of the entire keyframe set over the time axis. $\lambda$ is the balancing hyper-parameter.

\noindent
\textbf{Computing $s(\mathbf{Q},\mathbf{F}_t)$.} This involves a vision-language (VL) module to measure whether $\mathbf{F}_t$ contains information for answering $\mathbf{Q}$. Although the target MLLM itself $G(\cdot)$ can play the role, its high computational cost can bring a major burden. In practice, we choose a cheaper VL model (\textit{e.g.}, CLIP~\cite{radford2021learning} or BLIP ITM~\cite{li2022blip}) for replacement.

\noindent
\textbf{Estimating $c(\mathcal{I})$.} Measuring coverage is an open problem which is related to the homogeneity of data distribution. In mathematics, Ripley's $K$-function~\cite{ripley1976second} is a popular way to measure homogeneity. Given the timestamp set $\mathcal{I}$ and any search radius $r<T$, the $K$-function of $r$, denoted as $\hat{K}(r)$, is proportional to the number of $(t_i,t_j)$ pairs satisfying $|t_i-t_j|<r$. The distribution of $\mathcal{I}$ is considered homogeneous (\textit{i.e.}, covering the entire time axis) if $\hat{K}(r)$ is approximately proportional to $r^2$.

To adapt $K$-function to computing coverage (closely related but a bit different from homogeneity) as well as reducing computational overhead, we introduce bins with width $r$ and approximate $\mathbb{I}(|t_i-t_j|<r)$ as whether $t_i$ and $t_j$ fall into the same bin. We perform a recursive partition. At the first level, we set $2$ bins with the bin width being $T/2$, \textit{i.e.}, the time axis $[0,T)$ is partitioned into $2$ non-overlapping bins, $[0,T/2)$ and $[T/2,T)$. With the numbers of keyframes falling within these bins being $m_1$ and $m_2$, $c(\mathcal{I})$ adds a penalty term $|m_1-m_2|$ since an uneven distribution implies weak coverage in the bin with fewer keyframes. At the second level, each of $[0,T/2)$ and $[T/2,T)$ is further partitioned into two bins, and the same calculation continues. The recursion stops at the $L$-th level, where $L\leqslant\lceil\log_2M\rceil$ is a hyper-parameter.

\subsection{Adaptive Keyframe Sampling}
\label{method:optimization}

With the complex definition of $c(\mathcal{I})$, it is difficult to find a closed-form or accurate optimization for Eqn~\eqref{eqn:objective2}. This part discusses an approximation. Compared to the baseline that only relies on $s(\mathbf{Q},\mathbf{F}_t)$ scores, we name such methods timestamp-aware optimization for its ability to consider timestamps for better keyframe selection results.

We first discuss two special cases. (1) When $\lambda=0$ (\textit{i.e.}, coverage is neglected), Eqn~\eqref{eqn:objective2} is solved by simply selecting the top-$M$ frames of the largest scores. We name this strategy \textbf{TOP}, short for `top sampling'; as shown in Figure~\ref{fig:strategies}, in some cases, it results in all keyframes being located within a small range of time and the MLLM missing important information in other moments. (2) When $\lambda\rightarrow+\infty$ (\textit{i.e.}, coverage is strictly guaranteed), Eqn~\eqref{eqn:objective2} is solved by selecting the frame within the highest score in each bin as the keyframe (when the number of bins exceeds $M$, the champion frames with the highest scores are preserved). We name this strategy \textbf{BIN}, short for `binned sampling'. This situation further degenerates to the uniform sampling baseline~\cite{zhang2024video} if a dummy VL model is used for scoring (\textit{i.e.}, $s(\mathbf{Q},\mathbf{F}_t)$ is a constant over $t$). We name this strategy \textbf{UNI}, short for `uniform sampling'.

In other cases ($0<\lambda\ll+\infty$), we adopt a hierarchical optimization method that follows the definition of $c(\mathcal{I})$. At the first level, we determine how to allocate $M$ keyframes into two bins, $[0,T/2)$ and $[T/2,T)$. We recall the relevance scores of all frames, $s(\mathbf{Q},\mathbf{F}_t)$, and compute the average scores over all frames (denoted as $s_\mathrm{all}$) and over $M$ frames with the highest scores (denoted as $s_\mathrm{top}$). If there is only one keyframe to be selected, or $s_\mathrm{top}-s_\mathrm{all}$ surpasses a threshold, $s_\mathrm{thr}$, we believe that it is important to guarantee the top-scored frames to be sampled (\textit{i.e.}, maximizing the first term of Eqn~\eqref{eqn:objective2}), so the algorithm directly returns the top-$M$ frames as keyframes. Otherwise, we split the current bin into two sub-bins with the number of keyframes evenly allocated (\textit{i.e.}, maximizing the second term of Eqn~\eqref{eqn:objective2}), and then recursively call the above programs in the sub-bins. We name this strategy \textbf{ADA}, short for `adaptive sampling'. Note that the hyper-parameter $\lambda$ is not explicitly tuned; its role is replaced by $s_\mathrm{thr}$.

Figure~\ref{fig:split} uses an example to show how adaptive sampling (\textbf{ADA)} works. \textbf{ADA} is a compromise between the special cases, \textbf{TOP} and \textbf{BIN}. As we shall see in experiments (see Section~\ref{experiments:ablation}), \textbf{ADA} absorbs the advantages of \textbf{TOP} and \textbf{BIN} and achieves the best practice of video understanding.


\section{Experiments}
\label{sec:experiments}

\subsection{Experimental Setup and Details}
\label{experiments:details}

\noindent\textbf{Dataset and evaluation.}
We utilize the popular LMMs-Eval~\cite{zhang2024lmms} to evaluate the performance of \name. We use two popular benchmarks LongVideoBench~\cite{wu2024longvideobench}, and VideoMME~\cite{fu2024video}, for long video understanding. The length of videos in this dataset can exceed one hour, so the quality of keyframe selection plays a crucial role in visual understanding. We establish \name beyond three recent video-based MLLMs (see the next paragraph). We do not tune the parameters of these MLLMs, but only change the input frames into those selected by \name. To highlight the importance of keyframe selection, we do not use video subtitles to assist question answering. This setting also allows us to weaken the impact of the LLM's strength and maximally focus on visual understanding.




\noindent\textbf{Implementation details.}
We investigate three video-based MLLMs as our baseline, namely, Qwen2VL~\cite{wang2024qwen2}, LLaVA-OV~\cite{li2024llava}, and LLaVA-Video~\cite{zhang2024video}. LongVideoBench and VideoMME contain multi-choice questions; to answer these questions, these MLLMs followed a similar prompt involving the question (in text), video frames (as tokens), and options (in text). Specifically, as the strongest baseline, LLaVA-Video used SigLIP~\cite{zhai2023sigmoid} as its vision encoder and Qwen2-7B~\cite{yang2024qwen2} as its large language model. With capacities to process up to $32$ or $64$ video frames, these MLLMs offer basic abilities of video understanding, but they were built upon uniformly sampled keyframes and can miss important information.

To reduce computational costs, we sample the candidate frames from the raw video at $1$ frame per second. The prompt (in text) and each $t$-th frame (as image) are fed into the text and visual encoders of BLIP~\cite{li2022blip} to obtain $\mathbf{Q}$ and $\mathbf{F}_t$, after which $s(\mathbf{Q},\mathbf{F}_t)$ is computed via image-text matching (ITM), \textit{i.e.}, the similarity between $\mathbf{Q}$ and $\mathbf{F}_t$. One can also replace BLIP with other vision-language models (\textit{e.g.}, CLIP~\cite{radford2021learning}); see the ablation in Section~\ref{experiments:ablation}. We use \textbf{ADA} sampling unless otherwise specified.

\begin{table}[!t]
\small
\centering
\setlength{\tabcolsep}{0.13cm}
\caption{Video-based question answering accuracy (\%) of different approaches on LongVideoBench (LVB) val and VideoMME (V-MME). \name is applied upon three baseline approaches. \textbf{Frames} and \textbf{LLM} indicate the number of video frames fed into the MLLM and the number of parameters in the LLM part, respectively.}
\begin{tabular}{l|cc|c|c}
\toprule
{\textbf{Method}}  & \textbf{Frames} &\textbf{LLM} & \textbf{LVB val}& \textbf{V-MME} \\
\midrule
\rowcolor{lightgray}
\textit{GPT-4V~\cite{openai2023gpt4v}}            &\textit{256}  &--  &\textit{61.3}   &\textit{59.9} \\
\rowcolor{lightgray}
\textit{GPT-4o~\cite{openai2024gpt4o}}            &\textit{256}  &--  &\textit{66.7}   &\textit{71.9} \\
\rowcolor{lightgray}
\textit{Gemini-1.5-Flash~\cite{team2023gemini}}  &\textit{256}  &--  &\textit{61.6}   &\textit{70.3} \\
\rowcolor{lightgray}
\textit{Gemini-1.5-Pro~\cite{team2023gemini}}    &\textit{256}  &--  &\textit{64.0}   &\textit{75.0} \\
\midrule
VideoLLaVA~\cite{lin2023videollava}     &8   &7B    &39.1   &39.9 \\
MiniCPM-V 2.6~\cite{yao2024minicpm}  &64  &8B    &54.9   &60.9 \\
PLLaVA~\cite{xu2024pllava}        &32  &34B   &53.2   &- \\
VILA~\cite{lin2024vila}          &-  &40B   &-   &60.1 \\
\midrule
Qwen2-VL~\cite{wang2024qwen2}       &32  &7B    &55.5   &57.6 \\
\textbf{Qwen2-VL w/ \name} &32   &7B  & \textbf{60.5} & \textbf{59.9}\\
\midrule
LLaVA-OV~\cite{li2024llava}       &32  &7B    &54.8   &56.5 \\
\textbf{LLaVA-OV w/ \name} &32   &7B  & \textbf{59.3} & \textbf{58.4}\\
\midrule
LLaVA-Video~\cite{zhang2024video}  &64   &7B  &58.9   &64.4 \\
\textbf{LLaVA-Video w/ \name} &64   &7B  &\textbf{62.7}   &\textbf{65.3} \\
\bottomrule
\end{tabular}
\label{tab:comparison_sota}
\end{table}

\begin{figure*}[!t]
\vspace{-0.7cm}
\centering
\includegraphics[width=0.98\linewidth]{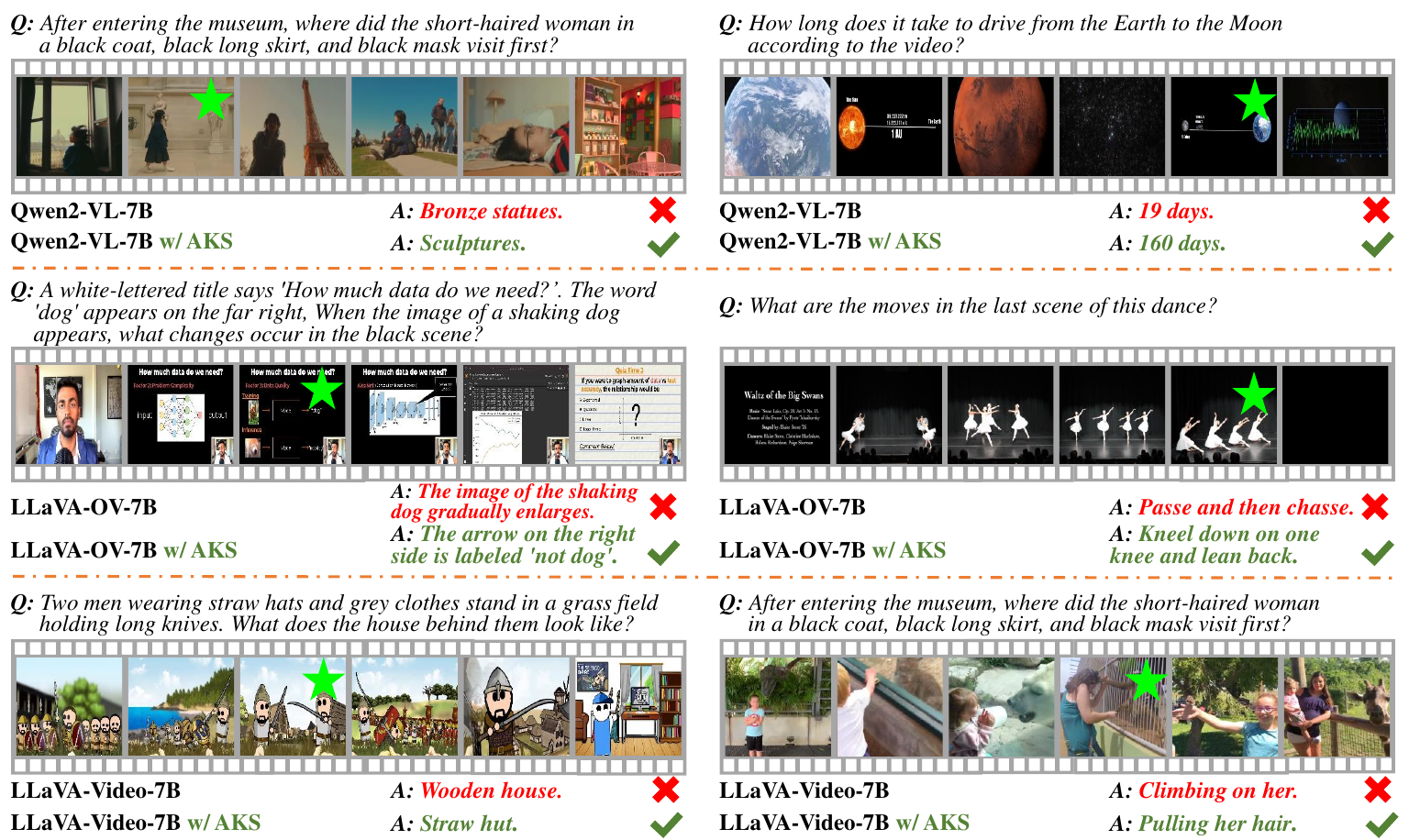}
\caption{\name improves the baseline MLLMs for video understanding. The left three examples come from LongVideoBench while the right three come from VideoMME. Green stars indicate keyframes selected by \name (note that $64$ keyframes are selected for each video).}
\label{fig:vis_comparison}
\end{figure*}

\subsection{Comparison to the State-of-the-Art}
\label{experiments:comparison}

\noindent\textbf{Quantitative results.} 
We first compare the accuracy of video question answering between our approach and some recent MLLMs. Results are summarized in Table~\ref{tab:comparison_sota}. \name brings consistent accuracy gain over three baselines, \textit{e.g.}, upon Qwen2VL, the improvement is $5.0\%$ on LongVideoBench and $2.3\%$ on VideoMME; even upon LLaVA-Video, the strongest baseline, these numbers are $3.8\%$ and $0.9\%$, respectively. These improvements not only make our method surpass other competitors with a similar computational complexity (\textit{i.e.}, input no more than $64$ frames, LLM no larger than 7B), but also allow it to achieve higher levels set by larger models (\textit{e.g.}, with \name, LLaVA-Video-7B reports $62.7\%$ on LongVideoBench, which is $0.8\%$ higher than the LLaVA-Video-72B model without \name, and $1.4\%$ and $1.1\%$ higher than GPT-4V and Gemini-1.5-Flash, two proprietary models using $256$ input frames).

\noindent\textbf{Qualitative results.}
In Figure~\ref{fig:vis_comparison}, we display representative video understanding results of \name (based on LLaVA-Video-7B) and others. One can see that the selected keyframes are closely related to the question; this allows the MLLM, with a limited capacity of context, to get a comprehensive view of question-related content and thus obtain the correct answer. As a side comment, we find that VideoMME contains many questions that require a high-level comprehension of the video content in which uniform sampling is a safe choice; nevertheless, \name still finds more informative frames and improves the accuracy, although the gain is smaller than that on LongVideoBench. Please also see the appendix for more examples.

\begin{figure*}[t]
\vspace{-0.8cm}
\centering
\includegraphics[width=1\linewidth]{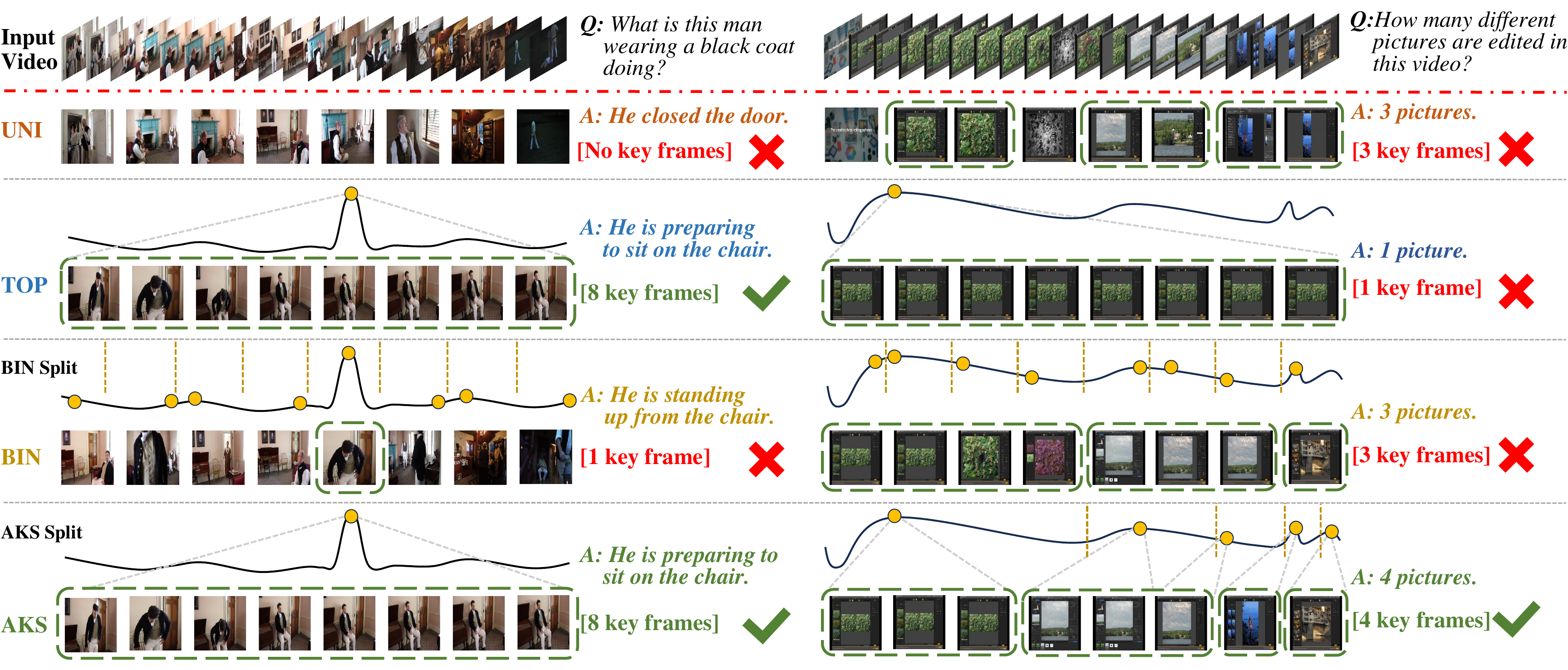}
\caption{Two examples of how different sampling strategies impact video understanding. The left case comes from LongVideoBench (focusing on one moment) and the right one comes from VideoMME (relying on multiple moments). Each curve shows the $s(\mathbf{Q},\mathbf{F}_t)$ score over time, and the yellow circles indicate the position of sampled keyframes. We also annotate the number of \textbf{true} keyframes and the reason for each failure case below the answer.}
\label{fig:strategies}
\end{figure*}

\subsection{Diagnostic on Keyframe Selection}
\label{experiments:diagnosis}

This part aims to diagnose how \name works and ablates design choices of \name. We build our test upon the strongest baseline, LLaVA-Video.

\begin{table}[!t]
\small
\centering
\setlength{\tabcolsep}{0.45cm}
\caption{Video-based question answering accuracy (\%) of different sampling strategies. LLaVA-Video-7B with \name is tested. Please refer to Section~\ref{method:optimization} for the explanations of these abbreviations and Section~\ref{experiments:diagnosis} for the analysis of results.}
\begin{tabular}{l|cccc}
\toprule
\textbf{Sampling}        & \textbf{LongVideoBench val}    & \textbf{VideoMME}   \\
\midrule
\textbf{UNI}     & 58.9          & 64.4           \\
\textbf{TOP}     & 62.4          & 63.7           \\
\textbf{BIN}     & 60.2          & 65.2  \\
\textbf{ADA}     & \textbf{62.7} & \textbf{65.3}  \\
\bottomrule
\end{tabular}
\label{tab:sampling_strategies}
\end{table}

\noindent\textbf{MLLMs benefit from better keyframes.}
To show how keyframe selection impacts video understanding, we test different strategies described in Section~\ref{method:optimization}. Table~\ref{tab:sampling_strategies} lists the results. Beyond the baseline (\textit{i.e.}, \textbf{UNI} sampling), \textbf{ADA} sampling (our default choice in Section~\ref{experiments:comparison}) achieves the best practice, while each of \textbf{TOP} and \textbf{BIN} samplings is better than the other in one benchmark. Note that the MLLM (\textit{i.e.}, LLaVA-Video-7B) remains unchanged throughout all these tests. In other words, all the improvements owe to \name in selecting higher-quality keyframes.

\noindent\textbf{Visualizing keyframe selection.}
Figure~\ref{fig:strategies} shows two representative examples and explains how the style of questions varies across LongVideoBench and VideoMME and how it impacts the preference between \textbf{TOP} and \textbf{BIN}. Many questions of LongVideoBench are focused on a simple moment (\textit{e.g.}, `\textit{What is a person doing at a specific time point?}'), so \textbf{TOP} sampling (\textit{i.e.}, without constraints in temporal distribution) often works well in locating these keyframes, while \textbf{BIN} sampling limits the number of keyframes within each bin and results in information loss. On the contrary, the questions of VideoMME often require the model to collect information from multiple moments (\textit{e.g.}, `\textit{How many times does something happen?}'), so \textbf{BIN} sampling is a safe choice to locate keyframes in different bins, while \textbf{TOP} sampling can lose information in weak peaks. \textbf{ADA} sampling absorbs the advantages of \textbf{TOP} and \textbf{BIN} strategies and adaptively allocates keyframes to the desired position (see the example in Figure~\ref{fig:split}) -- this is why it achieves the best results in both benchmarks.

\begin{figure}[!t]
\vspace{-0.7cm}
\centering
\includegraphics[width=1\linewidth]{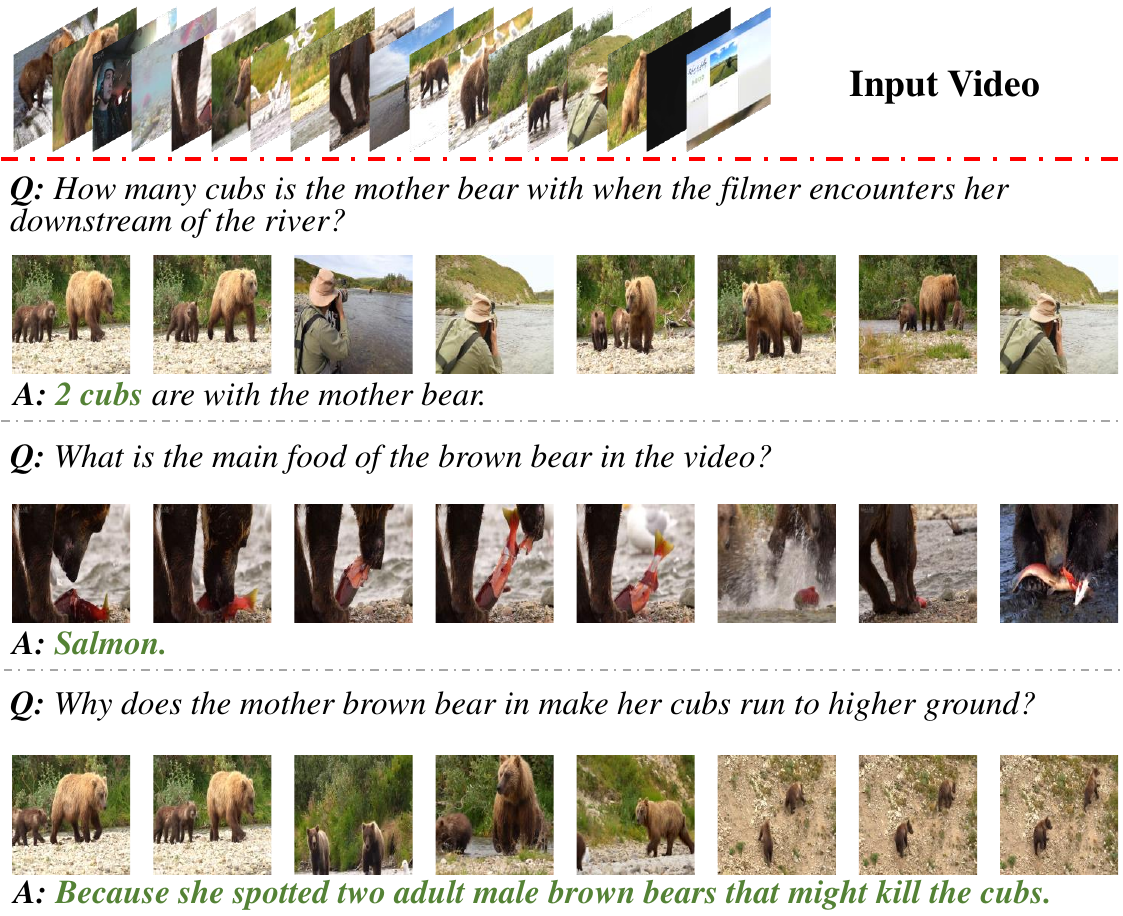}
\caption{\name selects different keyframe sets to answer different questions. All answers are correct.}
\label{fig:same_v_diff_q}
\end{figure}

Figure~\ref{fig:same_v_diff_q} shows an interesting example that, on the same input video, \name selects different sets of keyframes based on the prompt. This increases the flexibility that a frozen MLLM can adapt to different scenarios.

\subsection{Ablative Studies}
\label{experiments:ablation}

\begin{table}[!t]
\vspace{-0.5cm}
\small
\centering
\setlength{\tabcolsep}{0.33cm}
\caption{Question answering accuracy (\%) \textit{w.r.t.} different sampling frequencies. LLaVA-Video-7B is used as the MLLM.}
\begin{tabular}{l|ccccc}
\toprule
 \textbf{Frames}  & \multicolumn{5}{c}{\textbf{Sampling Frequency (fps)}} \\
\textbf{of MLLM}   & 1  & 0.5  & 0.25  & 0.125  & 0.1 \\
\midrule
\multicolumn{6}{l}{\textit{LongVideoBench val}} \\
16   & \textbf{61.6}   & 60.7  & 60.6  & 61.1  & 59.4\\
32   &         61.9   & \textbf{62.1}  & 59.8  & 60.2  & 58.5\\
64   & \textbf{62.7}   & 62.2  & 61.8  & 60.1  & 60.1\\
\midrule
\multicolumn{6}{l}{\textit{VideoMME }} \\
16   & 62.2  & \textbf{63.0}  & 62.2  & 61.0  & 61.6\\
32   & 64.6   & 64.7  & \textbf{65.1}  & 64.4  & 64.4\\
64   & \textbf{65.3}   & 65.1  & 64.9  & 64.0  & 64.2\\
\bottomrule
\end{tabular}
\label{tab:ranges}
\end{table}

\noindent\textbf{The frequency of sampling keyframe candidates.}
To decrease the extra computational cost brought by \name, we sample fewer keyframe candidates (\textit{i.e.}, one frame per $2$/$4$/$8$/$10$ seconds, and compare the results with the standard $1$-fps method. Results are summarized in Table~\ref{tab:ranges}. On LongVideoBench, even at $0.1$ fps, all results are higher than the baseline (\textit{i.e.}, $57.4\%$, $57.9\%$, $58.9\%$ at $16$, $32$, $64$ frames). VideoMME shows a similar trend, and $0.25$ fps seems a safe option to surpass the baseline. It is worth exploring more efficient pre-filtering algorithms towards a better tradeoff between accuracy and efficiency.


\begin{table}[!t]
\small
\centering
\setlength{\tabcolsep}{0.38cm}
\caption{Question answering accuracy (\%) \textit{w.r.t.} different VL scorers. LLaVA-Video-7B is used as the MLLM.}
\begin{tabular}{lcccc}
\toprule
\textbf{Frames} &\textbf{Uniform} & \textbf{BLIP} & \textbf{Sevila}  & \textbf{CLIP}\\
\midrule
\multicolumn{5}{l}{\textit{LongVideoBench val}} \\
16& 57.4& \textbf{61.6}& 59.2& 60.2\\
32& 57.9& \textbf{61.9}& 60.9& \textbf{61.9}\\
64& 58.9& \textbf{62.7}& 61.5& 62.2\\
\midrule
\multicolumn{5}{l}{\textit{VideoMME}} \\
16& 60.6& 62.2& 63.0& \textbf{63.1}\\
32& 63.9& 64.6& 63.7& \textbf{65.0}\\
64& 64.4 & 65.3 & 65.1& \textbf{65.6}\\
\bottomrule
\end{tabular}
\label{tab:scorers}
\vspace{-0.1cm}
\end{table}

\noindent\textbf{The VL model for computing $s(\mathbf{Q},\mathbf{F}_t)$ scores.}
We analyze the impact of using different VL models for computing prompt-frame relevance. We study three options, \textit{i.e.}, BLIP~\cite{li2022blip} (the default choice in this paper), Sevila (used in~\cite{yu2024self}), and CLIP~\cite{radford2021learning}. Results are summarized in Table~\ref{tab:scorers}. We find that BLIP works better on LongVideoBench while CLIP works better on VideoMME. This is because CLIP was trained on generic image-text pairs while BLIP learned from object-level data -- correspondingly, questions in LongVideoBench and VideoMME are more focused on objects and global perception, respectively.



\begin{table}[!t]
\footnotesize
\centering
\setlength{\tabcolsep}{0.055cm}
\caption{Ablating $L$ and $s_\mathrm{thr}$ together. Left: LVB, Right: V-MME.}
\begin{tabular}{ccccccc}
\toprule
\backslashbox{$L$}{$s_\mathrm{thr}$} &0.0 & 0.2 & 0.4 & 0.6 & 0.8 & 1.0 \\
\midrule
1 &62.4/63.8 &	62.4/64.0 &	62.5/64.2 &	62.0/64.1 &	61.8/63.8 &	61.9/64.0 \\
2 &62.4/63.8 &	62.0/64.0 &	62.4/64.0 &	61.8/63.5 &	61.7/63.4 &	62.0/63.6 \\
3 &62.4/63.8 &  \textbf{62.8}/64.0 &	62.6/54.5 &	62.1/64.4 &	62.2/64.4 &	62.1/64.4 \\
4 &62.4/63.8 &	62.7/64.1 &	62.7/64.3 &	62.2/64.9 &	62.1/65.0 &	62.2/65.0 \\
5 &62.4/63.8 &	62.7/64.1 &	62.2/64.7 &	61.7/65.0 &	61.3/\textbf{65.3} &	61.7/65.2 \\
6 &62.4/63.8 &	62.7/64.0 &	62.3/64.5 &	61.8/65.0 &	61.3/65.1 &	61.4/65.1 \\
\bottomrule
\end{tabular}
\label{tab:ablation on depth}
\vspace{-0.5cm}
\end{table}

\noindent\textbf{\textbf{ADA} hyper-parameters, $L$ and $s_\mathrm{thr}$.}
Lastly, we study the impact of $L$ and $s_\mathrm{thr}$. Results are summarized in Table~\ref{tab:ablation on depth}. One can see that LongVideoBench prefers smaller $L$ and $s_\mathrm{thr}$ values than VideoMME. This is because the key information on LongVideoBench is more concentrated (\textit{i.e.}, many questions are related to single moments) while that on VideoMME is more diverse (multi-moment data are required for question answering). \name offers a flexible ability to switch between different `modes' and achieves better results in both datasets.

\subsection{Generalization to Other Tasks}
\label{experiments:generalization}

\begin{figure}[t]
\vspace{-0.7cm}
\centering
\includegraphics[width=0.99\linewidth]{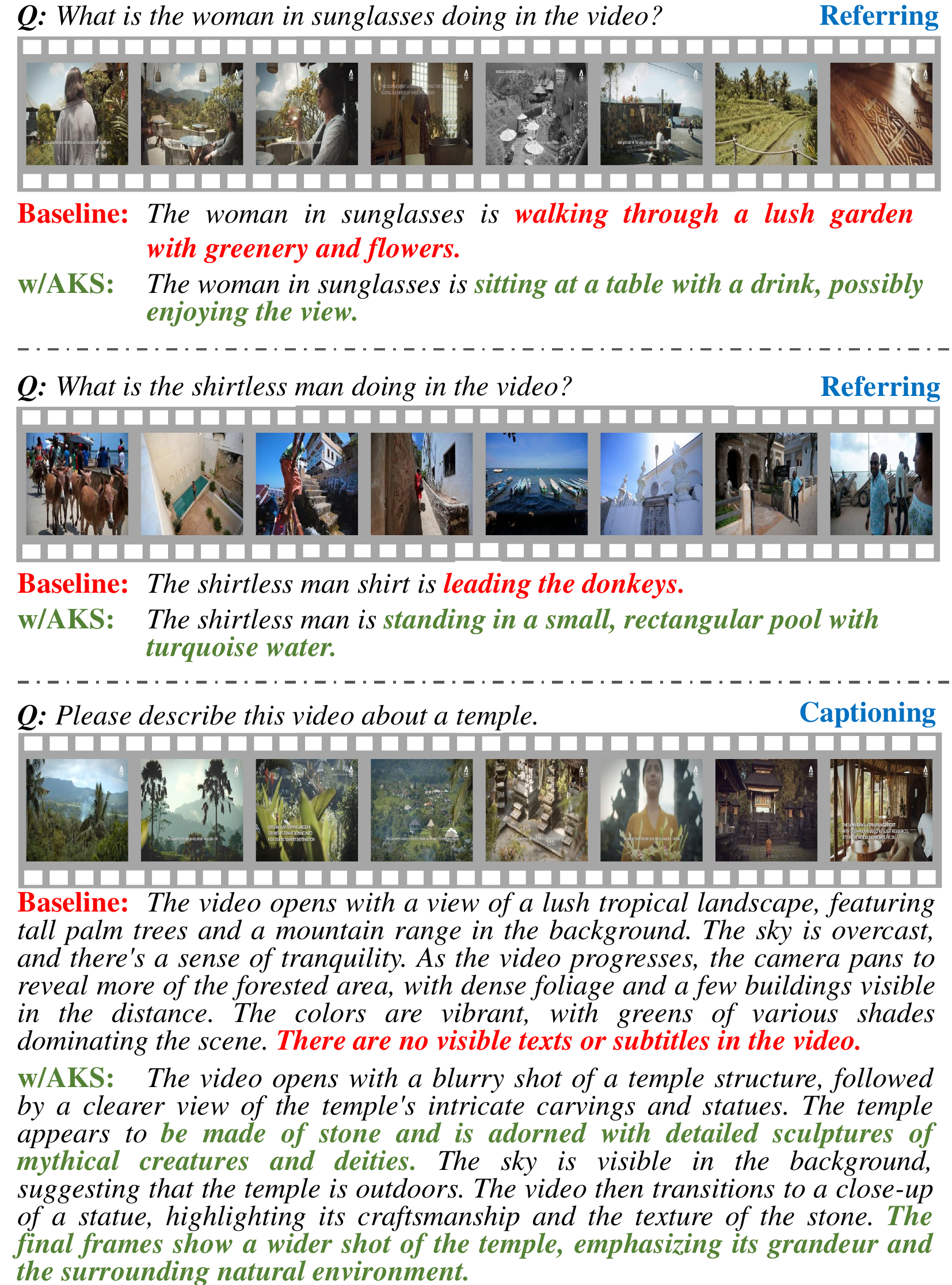}
\caption{Extending \name for video referring and captioning. The baseline results are produced upon uniform keyframe sampling. Red and green texts indicate incorrect and correct descriptions.}
\label{fig:generalization}
\vspace{-0.2cm}
\end{figure}

Being an off-the-shelf algorithm, \name is easily applied to other video understanding tasks. Here we showcase two examples known as video referring and captioning. For this purpose, we use the LLaVA-Video-7B model, switch the text prompt into `{What is [target] doing in the video?}' or `\textit{Please describe this video.}', and remove the options. Qualitative results are shown in Figure~\ref{fig:generalization}. As seen, to obtain a comprehensive description of long videos, it is crucial to locate keyframes and feed them into the MLLM as visual contexts. \name benefits from its keyframe selection ability and helps the MLLM to generate much better answers.

\section{Conclusions}
\label{sec:conclusions}

This paper focuses on improving the ability of MLLMs for long video understanding. The main difficulty arises from the limited capacity of MLLMs which urges us to feed informative visual tokens into the model. For this purpose, we present the \fullname (\name) algorithm which (1) uses a vision-language model to estimate the \textbf{relevance} and (2) applies an adaptive optimization algorithm to facilitate the \textbf{coverage} of selected keyframes. Quantitative and qualitative studies validate the effectiveness of \name over different baselines and benchmarks. Our work reveals that a pre-filtering stage brings considerable benefit to video understanding and advocates for further studies in this direction.

{
    \small
    \bibliographystyle{ieeenat_fullname}
    \bibliography{main}
}

\appendix
\setlength{\parskip}{0pt}
\section{Details of the ADA Algorithm}
In Algorithm~\ref{alg:algorithm}, we present the detailed pseudocode of our \textbf{ADA} algorithm. To accelerate the experimental process, we pre-process the video frames (sampled at 1 frame per second along with the corresponding questions) by inputting them into the VL scorer to obtain the corresponding scores.

\noindent
These scores are then stored in a list referred to as $matching\underline{~}score$. Each element in $matching\underline{~}score$ consists of the matching score for a specific video frame and the corresponding question. We begin by employing a recursive strategy to partition the matching\underline{~}scores list into sublists of varying lengths, according to the partitioning rule outlined in  Section~\textcolor{red}{3.3}. Subsequently, based on the lengths of these sublists, we select different numbers of frames with the highest matching scores from each sublist to construct the final set of video frames. This final set is then sent to the language model for visual understanding.

\begin{algorithm}[!t]
\caption{\textbf{ADA}: Adaptive Keyframe Selection}
\label{alg:algorithm}
\KwIn{$\mathrm{matching\_scores}$: A list, where each element is the matching score of a frame and the corresponding question \\
      $level$: Current recursion level \\
      $max\_level$: Maximum recursion level \\
      $s_\mathrm{thr}$: Threshold \\
      $M$: Number of frames to select}
\KwOut{$\mathrm{selected\_frames}$: Indices of the selected $M$ frames}

\SetKwFunction{SplitSegments}{SplitSegments}
\SetKwFunction{SelectFrames}{SelectFrames}
\SetKwInOut{Input}{Input}
\SetKwInOut{Output}{Output}

\BlankLine
\SetKwProg{Fn}{Function}{:}{}
\Fn{\SplitSegments{$\mathrm{matching\_scores}$, $level$, $max\_level$, $s_\mathrm{thr}$, $M$}}{
    $\mathrm{split\_scores} \leftarrow$ [] \tcp{List of completed segments}
    $\mathrm{new\_scores} \leftarrow$ [] \tcp{List of segments to further split}
    
    \ForEach{$matching\_score$ in $\mathrm{matching\_scores}$}{
        $s_\mathrm{all} \leftarrow \mathrm{mean}(matching\_score)$ \\
        $s_\mathrm{top} \leftarrow \mathrm{mean}(\text{topk}(matching\_score, M))$ \\
        $m \leftarrow s_\mathrm{top} - s_\mathrm{all}$ \\
        
        \If{$m \geq s_\mathrm{thr}$}{
            Append $matching\_score$ to $\mathrm{split\_scores}$
        }
        \ElseIf{$level < max\_level$}{
            Split $matching\_score$ into two bins from the center, denoted as $split_1$ and $split_2$ \\
            Append $split_1$ and $split_2$ to $\mathrm{new\_scores}$
        }
    }
    
    \If{$\mathrm{new\_scores}$ is not empty}{
        $\mathrm{deeper\_scores} \leftarrow$ \SplitSegments($\mathrm{new\_scores}$, $level+1$, $max\_level$, $s_\mathrm{thr}$, $M // 2^{level}$) \\
        $\mathrm{split\_scores} \leftarrow \mathrm{merge}(\mathrm{split\_scores}, \mathrm{deeper\_scores})$
    }
    
    \Return $\mathrm{split\_scores}$
}

\Fn{\SelectFrames{$segments$, $M$}}{
    $\mathrm{total\_length} \leftarrow$ Total length of all $segments$ \\
    $\mathrm{selected\_frames} \leftarrow$ [] 
    
    \ForEach{$segment$ in $segments$}{
        $m_i \leftarrow \lfloor M \times \text{length}(segment) / \mathrm{total\_length} \rfloor$ \\
        Select the top $m_i$ highest-scoring frame indices from $segment$ \\
        Append the selected indices to $\mathrm{selected\_frames}$
    }
    
    \Return $\mathrm{selected\_frames}$
}

\BlankLine
\SetKwProg{Main}{Main}{:}{}
\Main{}{
    $\mathrm{matching\_scores} \leftarrow$ [$matching\_score$] \\
    $segments \leftarrow$ \SplitSegments($\mathrm{matching\_scores}$, $level$, $max\_level$, $s_\mathrm{thr}$) \\
    $\mathrm{selected\_frames} \leftarrow$ \SelectFrames($segments$, $M$) \\
    \Return $\mathrm{selected\_frames}$
}
\end{algorithm}

\section{More Visualization Results}

\begin{figure*}[!t]
\centering
\includegraphics[width=0.99\linewidth]{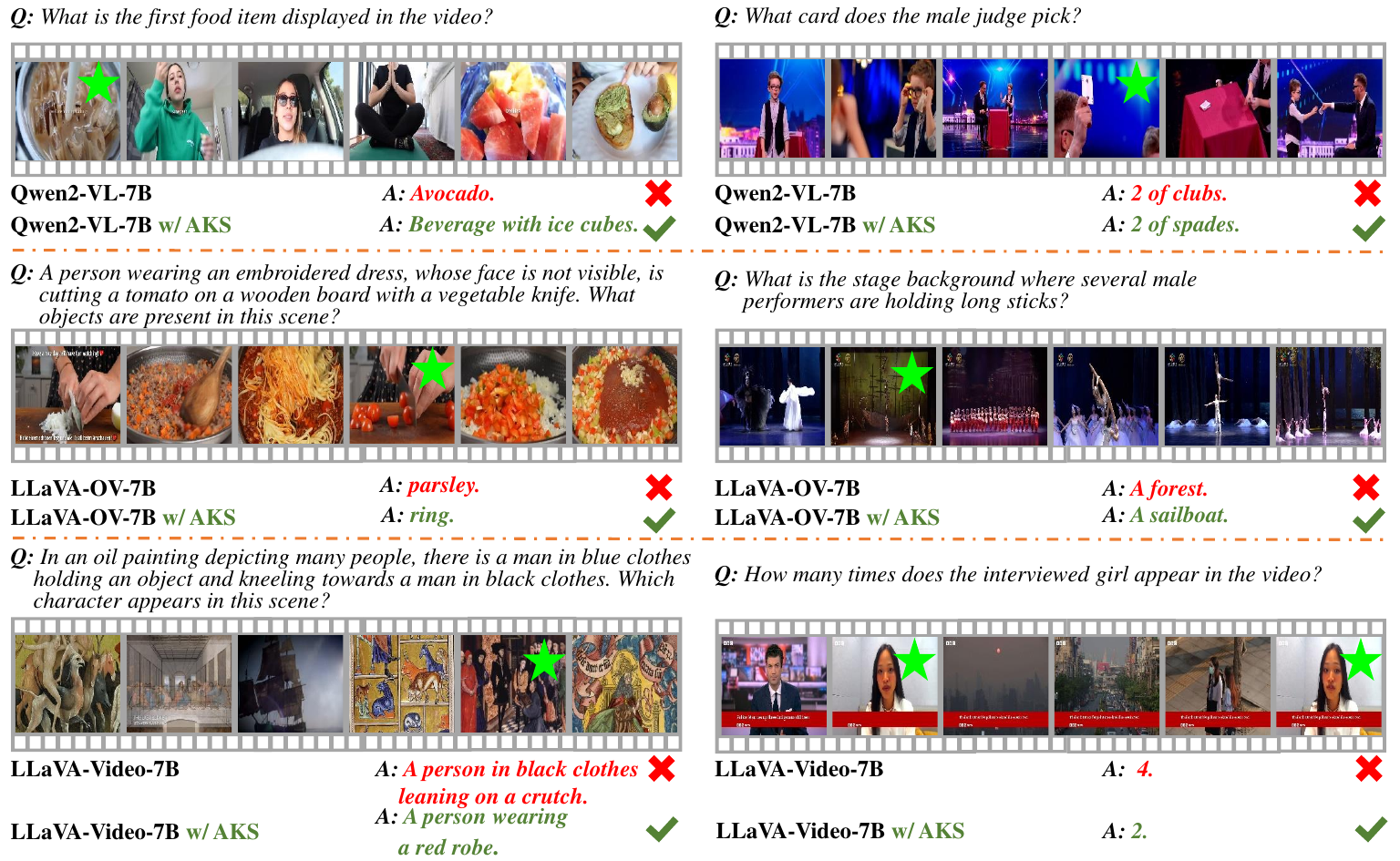}
\caption{More examples of AKS enhance the baseline MLLMs for video understanding. The left three examples come from LongVideoBench~\cite{wu2024longvideobench} while the right three are from VideoMME~\cite{fu2024video}. Green stars indicate keyframes selected by AKS.}
\label{fig:compare}
\end{figure*}

In Figure~\ref{fig:compare}, we show more examples of video understanding results of AKS (based on three baselines, LLaVA-Video-7B~\cite{zhang2024video}, Qwen2-VL-7B~\cite{wang2024qwen2}, and LLaVA-OV-7B~\cite{li2024llava}). As shown, our approach benefits from the ability to locate keyframes so that the MLLM receives effective visual information for understanding. The ability easily transfers to various MLLMs in a plug-and-play manner.

\end{document}